# TDIP: Tunable Deep Image Processing, a Real Time Melt Pool Monitoring Solution

Javid Akhavan
Mechanical Engineering
Stevens Institute of Technology
Hoboken, USA
Jakhavan@stevens.edu
0000-0002-6485-5986

Youmna Mahmoud
Mechanical Engineering
Stevens Institute of Technology
Hoboken NJ, USA
yelsayed@stevens.edu

Ke Xu
Mechanical Engineering
Stevens Institute of Technology
Hoboken NJ, USA
kxu7@stevens.edu
0000-0003-2439-1942

Jiaqi Lyu
Mechanical Engineering
Stevens Institute of Technology
Hoboken NJ, USA
jlyu2@stevens.edu
0000-0002-6259-1577

Souran Manoochehri
Mechanical Engineering
Stevens Institute of Technology
Hoboken NJ, USA
smanooch@stevens.edu
0000-0002-7189-6356

*Abstract*—In the era of Industry 4.0, Additive Manufacturing (AM), particularly metal AM, has emerged as a significant contributor due to its innovative and cost-effective approach to fabricate highly intricate geometries. Despite its potential, this industry still lacks real-time capable process monitoring algorithms. Recent advancements in this field suggest that Melt Pool (MP) signatures during the fabrication process contain crucial information about process dynamics and quality. To obtain this information, various sensory approaches, such as high-speed cameras-based vision modules are employed for online fabrication monitoring. However, many conventional in-depth analyses still cannot process all the recorded data simultaneously. Although conventional Image Processing (ImP) solutions provide a targeted tunable approach, they pose a trade-off between convergence certainty and convergence speed. As a result, conventional methods are not suitable for a dynamically changing application like MP monitoring. Therefore, this article proposes the implementation of a Tunable Deep Image Processing (TDIP) method to address the data-rich monitoring needs in real-time. The proposed model is first trained to replicate an ImP algorithm with tunable features and methodology. The TDIP model is then further improved to account for MP geometries and fabrication quality based on the vision input and process parameters. The TDIP model achieved over 94% estimation accuracy with more than 96% R2 score for quality, geometry, and MP signature estimation and isolation. The TDIP model can process 500 images per second, while conventional methods taking a few minutes per image. This significant processing time reduction enables the integration of vision-based monitoring in real-time for processes and quality estimation.

## I. Introduction

Metal additive manufacturing (AM) has gained significant attention and has been widely adopted in both industrial applications and academic research [1]. The two most commonly used types, that are laser based in nature, are laser directed energy deposition (DED) and powder bed fusion (PBF). These techniques offer a lot of advantages in manufacturing complex-shaped parts that are challenging to make using conventional methods, such as rolling, forging and casting [2]. Laser- directed energy deposition (L-DED), which is the application presented in this paper, is a type of DED that has been gaining a lot of attention in today's world, because it is relatively more cost effective, flexible and suitable for fabricating complex metallic geometries, with faster build times [3]. Known for its ability to produce intricate components with exceptional mechanical properties, L-DED has become a popular choice for a wide range of industries, including aerospace, defense, and medical. It is also noteworthy that L-DED is compatible with various materials, including metals, ceramics, and composites [4] But despite its numerous advantages, the DED process faces reliability challenges due to the multitude of parameters involved in the process, which may result in defects in the final product. L-DED involves the use of a gas flow to propel metal powder through a nozzle onto a substrate, while also subjecting it to a laser beam to achieve fusion. To better understand this complex physics of the underlying melt pool phenomenon, and ultimately optimize the process parameters to achieve accurate and reliable prints, the process-structure-property (PSP) relationship needs to be fully understood [5]. Literature has shown that single-track prints are often used in metal AM to investigate the impact of process parameters, such as laser power, laser speed, and powder feed rate, on the microstructure and physical characteristics of samples, as well as to determine the optimal conditions for producing larger builds, owing to their simplicity [4], [6]–[8].When energy and material fuse in the metal AM process, they create a molten metal pool known as the "melt pool" phenomenon. This is the first instance where key variables interact, and real-time monitoring can provide in-situ information about the build [9]. There has been a growing interest in melt pool monitoring owing to its significance and usefulness in predicting the quality of deposited material, fine-tuning parameters, designing control systems, and detecting defects.

Numerous researchers have then proposed the use of melt pool monitoring as a method for predicting process conditions by analyzing melting event signals, splatters, and/or signatures. For instance, researchers have monitored single-track prints using high-speed cameras [10], [11] and inline coherent imaging [12], to detect issues such as balling, continuous tracks, and keyhole porosity. Ye et al. [13] correlated captured and observed plume and splatter signatures to existing balling defects. Clijsters et al. [14] used a coaxial setup with a high speed NIR thermal CMOS camera and a photodiode to capture the melt pool area and intensity, for error detection purposes.

Using these online monitoring sensor data, researchers were also able to derive correlations between the metal AM process parameters (laser power, scanning speed, powder feed rate) and melt pool dimensions through observations and statistical methods. Colodrón et al. [15] used another coaxial CMOS camera to observe the melt pool and used an emissivity-based algorithm to measure its width. Araujo et al. [16] later improved upon this by utilizing an elliptical approximation, similar to the one used in this paper. Guijun et al. [17] examined the effect of process parameters on the temperature, width, and height of the melt pool. Their findings indicated that the laser power significantly impacts the track width. Sampson et al. [9] proposed a novel image processing algorithm providing a more precise width measurement of the melt pool. The proposed method surpasses the conventional emissivity-based edge detection techniques, exhibiting less than 1% deviation error.

This utilization of camera-based methods can generate enormous amounts of data, which can be efficiently and rapidly processed through machine learning (ML). This statistical-based approach has the ability to learn feature selection and extraction independently, making it a highly useful tool in studying AM processes. Several facets of DED, including melt pool monitoring, process parameter optimization, and defect identification, have been tackled with machine learning algorithms. For instance, Jeon et al. [18] has unveiled a revolutionary method that combines an artificial neural network, a co-axial infrared camera, and a laser line scanner to determine the depth of melt pools during directed energy deposition additive manufacturing. The authors' technique has a mean absolute percentage error of 8.95%, which is an improvement over earlier techniques. Pandiyan et al. [19] introduced an innovative method for assessing the effectiveness of DED procedures, in which co-axial process zone imaging and deep contrastive learning are used to evaluate the collected pictures and identify various types of defects. Deep contrastive learning is specifically used to automatically categorize and identify holes, balling, and non-homogeneous microstructure forms in the metal being constructed. It has been shown that this method can help improve the effectiveness and precision of quality control in DED processes. Lim et al. used machine learning techniques to identify the best manufacturing settings for the directed energy deposition (DED) process [20]. By conducting and assessing titanium-alloy single track experiments, the authors derived relationships between the process parameters, namely, laser power and scan speed, and the deposited surface color. By using random forest and support vector machine multi-classification models, authors were able to select the best surface color based on the structural and mechanical properties of the prints. This study is a prime example of the value of machine learning in increasing the productivity and quality of additive manufacturing processes. A novel software framework for in-situ monitoring of Laser Directed Energy Deposition through Powder Feeding (LDED-PF) based Metal Additive Manufacturing is described in a recent study by Farzaneh Kaji [21]. By employing a laser scanner sensor to scan the part's surface and machine learning algorithms to analyze the 3D model, this platform uses machine learning approaches to find surface irregularities. According to the researchers, this technique could improve the precision and quality of LDED-PF components, which is important in fields like aerospace and medical devices that demand great accuracy.

Recent studies have implemented machine learning methods to perform a real-time prediction of the melt-pool geometry. Zhu et al. [22] presented three optimized machine models such including an XGBoost model scoring at least 92% accuracy in predicting the melt pool height, width and depth. However, it is only efficient at predicting the extreme values ~ maximum and minimum values, and is weak in predicting intermediate values. Kim et al. [23] showed the importance of the spatter reduction tool for accurate prediction models when collecting melt pool data during laser powder bed fusion. They used a stacked convolutional denoising autoencoder (SCDAE), but an extended loss function may be required to keep the melt pool intact while removing spatters, depending on the training dataset. They also proposed the Kalman filter approach to address the measurement uncertainty in position estimation of the melt pool images. Computation speed remains a challenge when performing real-time control. Yang et al. [24] relied on down-sampling to reduce the processing time, while their predictive accuracy is still under 90%.

This paper details a deep learning model for analyzing Laser Directed Energy Deposition (LDED) AM processes in real-time. The implemented model compared to the conventional methods discussed is appropriate for industrial applications that need swift online decision-making and modifications since it can analyze and estimate the fabrication quality, with a faster processing turnaround. Due to the tunable capability developed, the model can easily adopt to the user expectations without any further need for retraining. The suggested strategy has the ability to cut down manufacturing expenses and time while simultaneously analyzing overall product quality.

## II. METHODOLOGY

In this study, conventional methods along with the state-of-the-art deep learning model solutions are implemented and discussed. To test out and illustrate the developed model's capabilities, a database of real time co-axial observations along with their cross-cut results published open access in [25] is used. This dataset contains data of various single track DED prints fabricated with various process parameters. For each track, the fabrication parameters, co-axial images recorded in real time and post-processed cross-sectional images and resulting quality information are provided. The diversity of fabrication strategies allows a better generalizable training for this study. Therefore, in each step, a randomized sample of the whole dataset is taken to make sure a balanced input from diverse settings is gathered.

### A. Image Processing (ImP)

The noise presence in the raw data, such as unwanted reflections and sparks, can hinder the clear Melt Pool (MP) signature identification from the laser signatures in Metal AM. To overcome this challenge, it is essential to detect and eliminate the noise from the raw image input. This process of noise elimination can help to emphasize the MP signatures and isolate meaningful information.

There are various possible solutions to this issue, including conventional ImP techniques such as OpenCV libraries, and state-of-the-art methods such as Deep Learning (DL). In this section, we implement and compare both methods with the objective of achieving real-time ImP capabilities.

*1) Open CV*

Open CV was utilized to execute the pre-processing stage in order to eliminate unwanted reflections and sparks from the raw input, while emphasizing the MP observation and minimizing laser signatures. To achieve this, the deviation map noise reduction method proposed by Akhavan et al. [25] was implemented in this study.

Various techniques can be employed to generate deviation maps, including the use of a local maxima filter, followed by a comparison with the raw data. In this study, a custom local maxima filter was applied over the raw input to generate the deviation map, which incorporates a tunable kernel size denoted as "F_size". Depending on the integration approach, the deviation map provides a method for minimizing or intensifying the co-axial camera's observed noise.

The scaling of the deviation map involves utilizing a tunable ratio labeled as "M_ratio", before being integrated into the raw data. The thresholding of the resulting data, utilizing "T_ratio", yields an isolated MP representation. The scaling's magnitude or the algorithms used to implement the deviation map can amplify different aspects of the MP such as outer or inner boundaries. Finally, any closed regions are filled, and only the largest connected region is retained as the MP. The implemented method's steps with their corresponding results are depicted in Fig. 1.

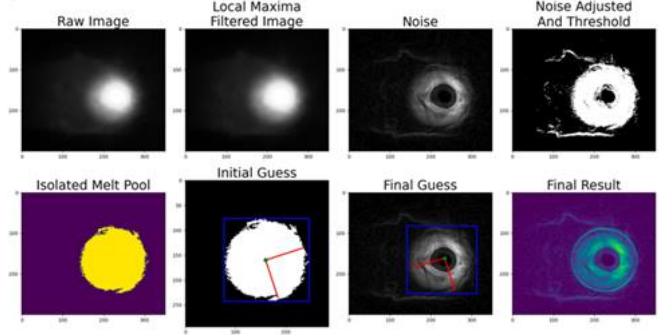

Fig. 1. Image processing steps taken to isolate melt pool signatures and dimensions.

After hyper-parameter fine-tuning using [26] the OpenCV method produced an acceptable MP isolation result, although it took a relatively long time (an average of 8 seconds/frame) to process each frame. For real-time applications, where the vision sensors' framerates range from tens to thousands of frames per second, conventional methods like the one mentioned above are not suitable. Therefore, there is a need for a real time capable algorithm to perform real time ImP while the fabrication is ongoing. To address this need, it is necessary to create a much faster processing algorithm capable of processing multiple frames in a fraction of a second.

*2) Deep Model*

Utilizing GPU-based processing for image data significantly improves yield time, which is necessary for real-time analysis [27]. Therefore, the implementation of Neural Networks over GPUs for ImP tasks has become a common practice in the field. However, although common Neural Network models outperform conventional Computer Vision methods, they lack the tunability that conventional methods provide. To better illustrate, an NN trained to estimate MP signatures from Gaussian smoothed data cannot process inputs with the Bilateral method and vice versa. Thus, the NN's applications are limited to the specific way the training task was performed. Adding tunable capability to the NN model architecture would allow for deeper and more generalized integration. In this article, a solution to this limitation is proposed, which eliminates NN limitations and expands its capacity to have tunable capability. Therefore, a Tunable Deep Neural Network-based ImP model is implemented in this article.

## III. TUNABLE DEEP IMAGE PROCESSING (TDIP) MODEL

### A. TDIP V.1

The TDIP model structure was developed with two main objectives in mind. Firstly, the model must meet the real-time processing criteria of processing more than 500 images per second. Secondly, the model should have the tunability of a conventional ImP algorithm.

To meet the first criterion, the model architecture depicted in Fig. 2 was designed. The TDIP model was built upon a Hybrid Auto Encoder-decoder model as described in [28]. This model architecture, with its various wide and deep layers, allows for a multi-aspect study of input images. The auto encoder decoder branch enables the bypassing of all process noise and reflections using image reconstruction techniques.

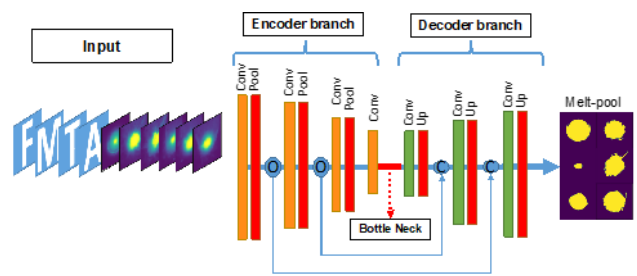

Fig. 2. TDIP V.1 model representation.

To fulfill the second criterion, the model must be customizable through user input, allowing for various Imp approaches. As a result, the model should take hyper-parameters as input, such as F_size, M_ratio, T_ratio, and A_mode, which significantly impact ImP results. To incorporate these inputs, the hint map information technique is utilized. This method involves creating an input hint map of the same size as the input image for each hyper-parameter needed to pass to the model. For each hyper-parameter (Hp), a layer of size D1*D2 is created and prefilled with the corresponding Hp value. Therefore, for N hyper-parameters, a tensor of shape N*D1*D2 is created. This information map is then appended to the input before model execution, allowing for regional hyper-parameter definition.

The TDIP model was trained using the Adam optimizer with the Mean Squared Error (MSE) loss metric and a dynamically changing learning rate in the range of [0.001 to 0.00001], and batch sizes varying from [50 to 5]. To build the training dataset, 2000 co-axial images were randomly sampled, and corresponding processed images were generated using randomly selected hyperparameters and OpenCV. The hyperparameters' ranges used during training are summarized in Table 1. The use of randomized hyperparameter selection allowed the model to learn the effects of each parameter on the Imp algorithm.

TABLE 1 HYPER-PARAMETER SELECTION RANGE USED FOR MODEL TRAINING.

| Hyper-parameter | Range |
|---|---|
| F_size | 3 – 5 – 7 – 9 |
| M_ratio | (0 10] |
| T_ratio | A_mode == 2: [0.005, 0.05] * M_ratio |
| | A_mode ≠ 2: [0.2, 0.8] |
| A_mode | 0: Deviation Map Addition<br>1: Deviation Map Subtraction<br>2: Deviation Map Alone |

Once the labels are generated from the base ImP model, each data point and its corresponding label were augmented and multiplied 168 times using image augmentation. Image augmentation summary is summarized in Table 2.

TABLE 2 AUGMENTATION SUMMARY FOR MODEL TRAINING

| Augmentation Method | Value Range | Resulting Multiplication |
|---|---|---|
| Rotation | 0 – 90 – 180 – 270 | 4 |
| Mirror | Horizontal – Vertical – diagonal (×2) | 4 |
| Shifts | ±20 pixels with 2-pixel steps | 21 |
| Total | | 168 |

*1) Melt Pool Geometry analysis*

To take it a step further, once the MP signatures have been identified and isolated, it becomes possible to study the MP geometries. In this study, the melt pool geometry is approximated using an ellipse model. This research work can be useful in fitting the geometric parameters of the Goldak's heat source distribution, that is commonly used in modeling the melt pool's heat flux as a moving heat source [29]. However, most of the proposed methods in literature rely on conventional computer vision algorithms such as Grid Search, Hough transformation, RANSAC, and others, which are computationally intensive. These methods blindly search the possible parameter space, looking for the best possible estimation. On average, depending on the input resolution, these methods can converge within a few seconds to minutes per each input frame. Obviously, such turnaround time would not be feasible for real-time processing and control applications. Therefore, in this section, a novel deep learning-based method is implemented to process the MP signatures in real-time and estimate the required geometries.

*2) Conventional Method*

To train a deep learning model, a comprehensive database of input/output pairs is necessary to ensure that the model learns all possible scenarios and becomes a generalizable estimator. To create inputs for the model, the database [25] can be processed and denoised with the TDIP model from the previous step. However, to generate a reliable and accurate label describing the ellipse geometries, a label generation method is required. Therefore, a grid search method is implemented using the technique proposed in [29]. The Grid Search method guarantees convergence to an acceptable result since it iterates over all possible solutions. Although this method is time-intensive, the results are reliable. On average, the Grid Search took 40 seconds per frame to achieve a reasonable within +-1 pixel accuracy result.

*B. TDIP V.2*

In order to utilize deep learning to estimate geometry, a regression model must be implemented. To improve efficiency and reuse the feature maps generated in the previous step through the TDIP model, a regression branch was integrated into the same network, enabling transfer learning for the estimation task. This approach, which was the central concept of the HCAE model presented in [30], facilitates faster model training by reusing the feature maps produced during the denoising process of the raw data. This approach enables the network to directly estimate geometries from the feature maps located in the bottleneck layer instead of utilizing the denoised data. Fig. 3 illustrates the updated TDIP model.

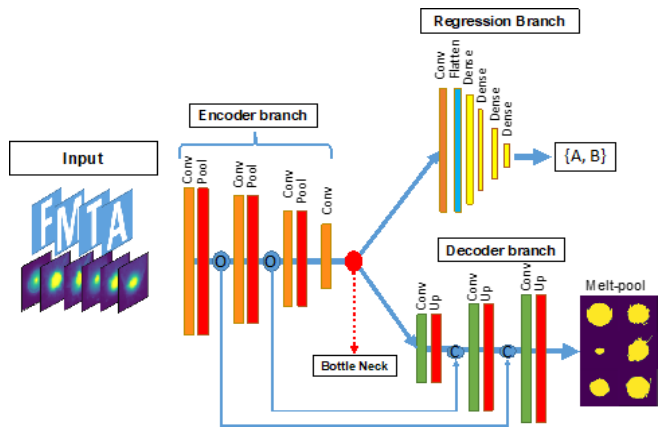

Fig. 3. TDIP V.2 model representation with melt pool geometry estimation branch.

The training of the TDIP V.2 model involved the use of the Adam optimizer and MSE loss for both the ImP and estimator branches. Multiple training strategies were employed throughout the process. Initially, the auto encoder decoder branch was trained on MP signature references. Then, the estimator branch with a dynamically changing loss weight was introduced. The loss weight of the estimator increased over time, while the loss weight of the AE branch decreased until convergence. To fine-tune the resulting network and ensure convergence, different learning rates and batch sizes were experimented with during the training process.

*1) Quality Estimation*

AM processes are subject to various dynamically changing environmental instabilities, such as machine degradation and inaccuracies. Even slight deviations in the earlier stages can lead to catastrophic quality issues in the final product. Therefore, early anomaly detection and quality estimation are essential for reliable manufacturing. To achieve such capabilities, real-time quality measurement is necessary. Previous work has demonstrated various functional quality assessment approaches to estimate the resulting fabrication quality based on observed process dynamics. For example, in single track printing, it is possible to determine how the printed track will form and what the final height and width will be by knowing the MP geometries. Many researchers have employed post-processing techniques, such as cross cuts, to measure laser penetration and MP depth. Lyu et al. [conference paper] proposed several machine learning solutions for estimating a track's final geometry using an ImP baseline. Their method showed that machine learning algorithms are capable of estimating Width (W), Height (H), and Depth (D) by analyzing the average MP's longest and shortest diagonal. However, due to the reliance on the average MP's geometries aggregated by conventional methods, this approach would not be suitable for real-time analysis. This study implements a deep model to address this gap, which can estimate H, D, and W per frame in real-time with an acceptable accuracy.

*C. TDIP V.3*

To perform geometrical analysis, some of the system's parameters such as laser power referred to as "Power", printer's speed, referred to as "Speed", and powder feed rate referred to as "Feed Rate" are necessary to be known by the model. Therefore, using the hint map information strategy discussed earlier, these parameters are stacked upon the input frames to inform the model of the printing dynamic. To reuse the already established feature maps, generated from the TDIP V.2 model, a third estimator branch is developed and attached to the bottle neck. This way, the model not only learns how to denoise the input observations but also calculates the MP geometries and resulting track fabrication qualities.

To train the TDIP V.3 model depicted in Fig. 4, not only the MP signatures and geometries are needed, but also quality observations to describe resulting fabrication geometries and quality is needed. Therefore, referring to the database [25], for each set of images, their corresponding H, D, and W reading from cross-cut microscope study are aggregated. Inputs to TDIP V.3 model are raw co-axial observations stacked over three process parameters and four ImP hyper parameters. Outputs expected from this model are: 1) denoised and isolated MP signatures, 2) MP' geometry estimations {A, B}, and 3) Resulting print geometry {H, D, W}.

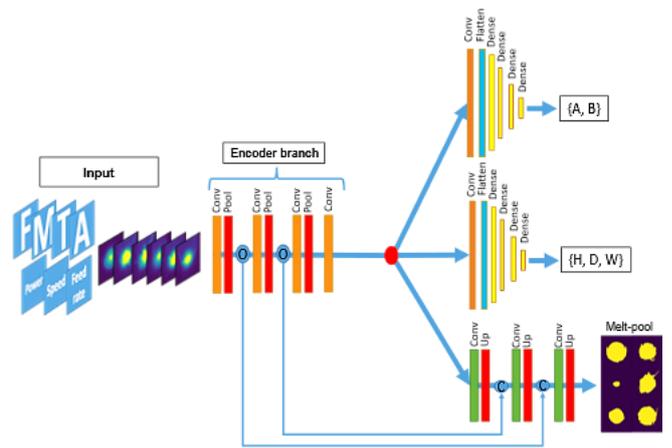

Fig. 4. TDIP V.3 model representation.

To train TDIP V.3, Adam optimizer with MSE loss function is used. Throughout the training process, similar to the previous model, one-by-one the output branches are introduced to the network with dynamically varying weights. The weight selection was performed manually based on the model's training convergence. In summary, first the AE branch is trained followed by MP geometry estimator. Once fully converged, the third branch, quality estimator is introduced. While training the weight between the three branches are changed ending with fully focused on the quality estimator branch. Throughout the training, learning rate and batch sizes are also dynamically updated to direct the model towards the optimal state.

## IV. RESULTS

This section summarizes all the TDIP model variation performances on the database [25] measured against the labels generated with conventional ImP methods. To start, first TDIP V.1 with only the ImP branch is discussed. Then, TDIP V.2 with the MP geometry estimator extension is evaluated. Finally, TDIP V.3 performance results against the conventional methods is measured. To further extend the TDIP V.3 performance measurement, the model outputs are compared against the point cloud data aggregated in real time for the same database.

*A. TDIP V.1 Results*

To measure the reliability and convergence of the TDIP V.1 model, R2 score metric was used. This metric shows how close the estimated results are from the expected values calculated by the conventional method based on OpenCV packages. To perform the measurement, 1000 samples with random processing hyper parameters are randomly selected. The selected data were compared with the training dataset, in case of similarity, a new data is sampled.

TDIP V.1 model achieved a 97% R2 score value which is proving the model capability to estimate significantly close to the ground truth labels derived by the conventional ImP method. Therefore, TDIP V.1, can efficiently replace the conventional method. To measure the processing capabilities, TDIP V.1 could process up to ~600 frames/second which compared to the conventional method with ~8 seconds/frame, it is a ~4800 times faster processing turnaround.

## B. TDIP V.2 Results

TDIP V.2, by having a second output branch, was developed to simplify the processing steps into a single model with multiple analysis capabilities. Depending on the input hyper-parameters, MP geometry readings are also influenced. Therefore, to create a reasonable testbed, another 1000 images with random hyper-parameters randomly were selected. Using the conventional methods mentioned earlier, they were processed to achieve the finest results. This process, specifically the grid search took close to three days. Using the TDIP V.2 model, the test set was processed and resulting MP signatures and geometries were recorded. TDIP V.2 took approximately ~2 seconds to process the test set which is ~120,000 times faster. To measure the estimation accuracy, first the MP signatures are compared with their ground truth using R2 metric. TDIP V.2 achieved 96% R2 value which is similar the previous version. To measure the geometry estimation accuracy, R2 metrics and 5pixel deviation tolerance metrics are used. TDIP V.2 estimated the MP geometries with 99.5%, and 99% R2 values and 94%, and 93% with 5pixel deviation tolerance metric for longest and shortest diagonals. TDIP V.2 estimation accuracy and its significant time efficiency makes it an even greater fit for real-time process monitoring.

## C. TDIP V.3 Results

Having TDIP V.1 and V.2, the benefits of deep learning technique for ImP and analysis became evident. To step even further, TDIP V.3 was introduced with the goal of non-destructive measurement capability. Just by looking at the co-axial data and knowing the process parameters, we needed a model capable of estimating the fabrication quality. Therefore, using the cross-cut section of database [25], a dataset of input/cross-cut readings was created. Using random images per each track with randomized hyper-parameters the TDIP V.3 model performance was tested. Using the same mentioned criteria of R2 for images and both R2 and 5pixel deviation metrics, the model outputs were compared. TDIP V.3 performance for ImP branch achieved 92% R2 score, and for MP geometry estimation R2 values of 98% and 97.5%, using 5pixel tolerance metric values of 88% and 86% was recorded for longest and shortest diagonal respectively. For the main goal, which was quality estimation, TDIP V.3 model achieved 97%, 96%, and 99% R2 values for H, W, and D features respectively. The 5-pixel metric for H, W, and D was 96%, 94%, and 98% respectively. A summary of all the performance values is provided in Table 3.

TABLE 3 SUMMARY OF TDIP PERFORMANCES AGAINST CONVENTIONAL MODELS.

| Model | Frames/second | Branch | R2 (%) | 5Pixel (%) |
|---|---|---|---|---|
| TDIP V.1 | ~600 | MP_Signatures | 97 | - |
| TDIP V.2 | ~500 | MP_Signatures | 96 | - |
| | | Geometry: A | 99.5 | 94 |
| | | Geometry: B | 99 | 93 |
| TDIP V.3 | ~450 all branches ~700 w/ only quality branch | MP_Signatures | 92 | - |
| | | Geometry: A | 98 | 88 |
| | | Geometry: B | 97.5 | 86 |
| | | Quality: H | 97 | 96 |
| | | Quality: D | 99 | 98 |
| | | Quality: W | 96 | 94 |

TDIP V.3 showed a great measurement capability for fabrication quality with above 96% R2 score. Although a very slight estimation accuracy drop was observed for geometry estimation and MP isolation, the model capability in real time evaluation of a destructive offline measurement. TDIP V.3 model achieved ~450frames/second turnaround. By only focusing on the quality estimation branch and deactivating the other two, the model can process ~700frames/second. The sample output illustration of TDIP V.3 for MP ImP is depicted in Fig. 5 for four randomly selected co-axial images. Various ImP hyper-parameter sets are randomly picked and applied. The OpenCV model MP outputs and TDIP model MP outputs are depicted to illustrate the TDIP model capability. As it was observed, in cases like (D, E), the OpenCV model has failed to achieve a reasonable MP, while the TDIP model, had a significantly better performance in MP signature isolation. In various cases such as rows (F, I, J, K) the OpenCV model failed to completely eliminate noise and unwanted segments and carried unwanted error data points along with the processed MP signature, while the TDIP model successfully eliminated and isolated the MP.

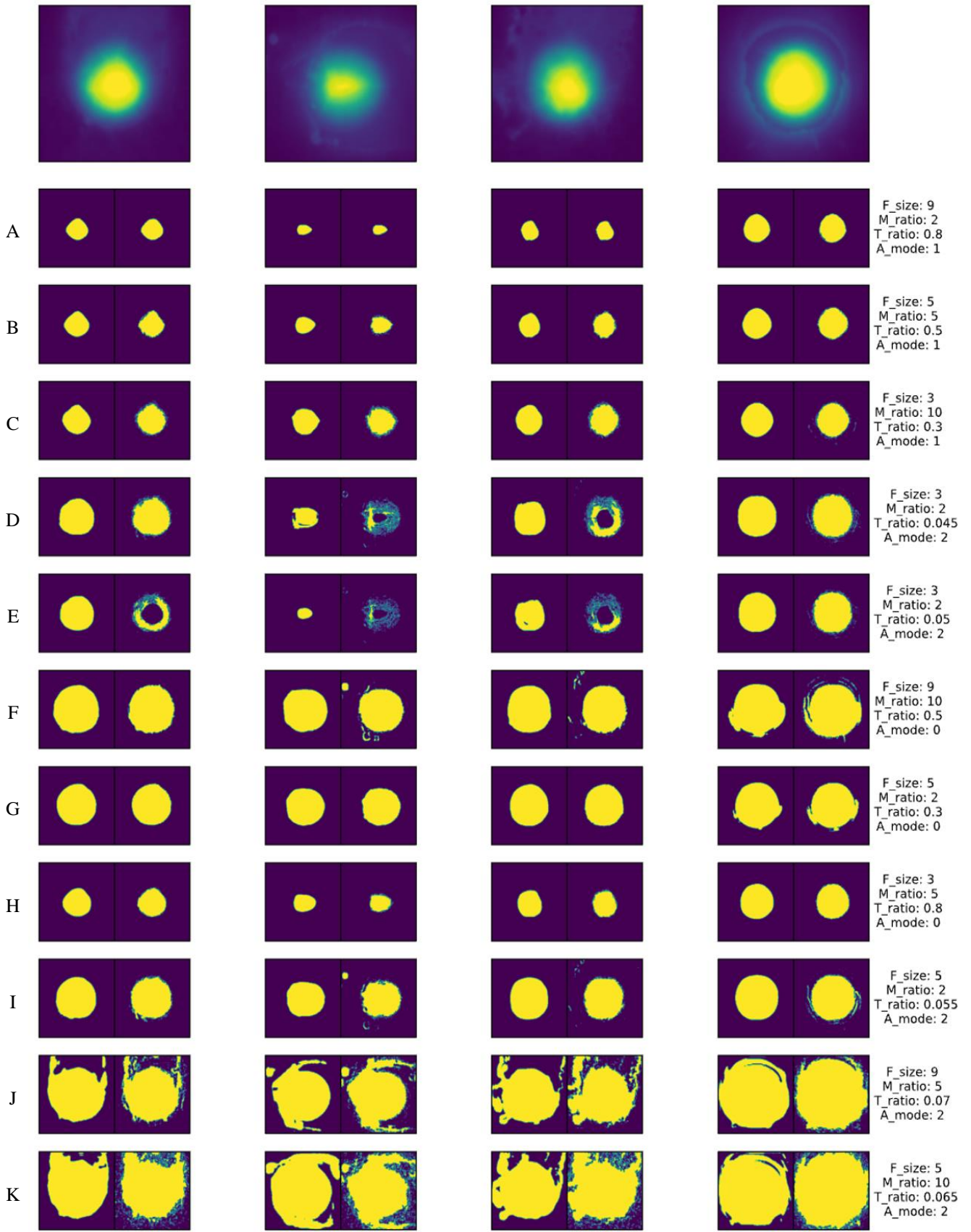

Fig. 5. TDIP Image Processing sample output with various hyper parameter sets. Top row is the raw input, The Left column is the TDIP output, The Right column is the OpenCV output.

To illustrate the TDIP model capability for quality estimation, for an unseen single-track print, Height, Depth, and Width values for three samples are estimated and depicted in Fig. 6. In this figure, the blue curve is the real time model output, red line is the smoothed model output using gaussian average, and dotted line is the actual measured values from experimental cross-cut data. Each single line was printed with different set of parameters leading to different printing conditions. As it was expected, the real time data follows the cross-cut values closely. It is worth mentioning that, although the cross-cut information is sparse and not continues, the TDIP model could capture the quality trends.

To account for printer's transition and stability, the dataset's first and last 15% are discarded. In the beginning and ending of each track segment, the machine undergoes a parameter shift phase which causes a slight DED laser's movement to pause, leading to a bulge print. TDIP model, although was not trained to understand this situation could perfectly estimate these unsteady samples.

## V. CONCLUSION

To conclude, using TDIP model structure, this research could propose a significant data processing improvement for co-axial image data flow. Observations with less than ~450frams/second can be analyzed. TDIP model also provides a tunability feature that allows adjustments be made over the algorithm and hyper-parameters.

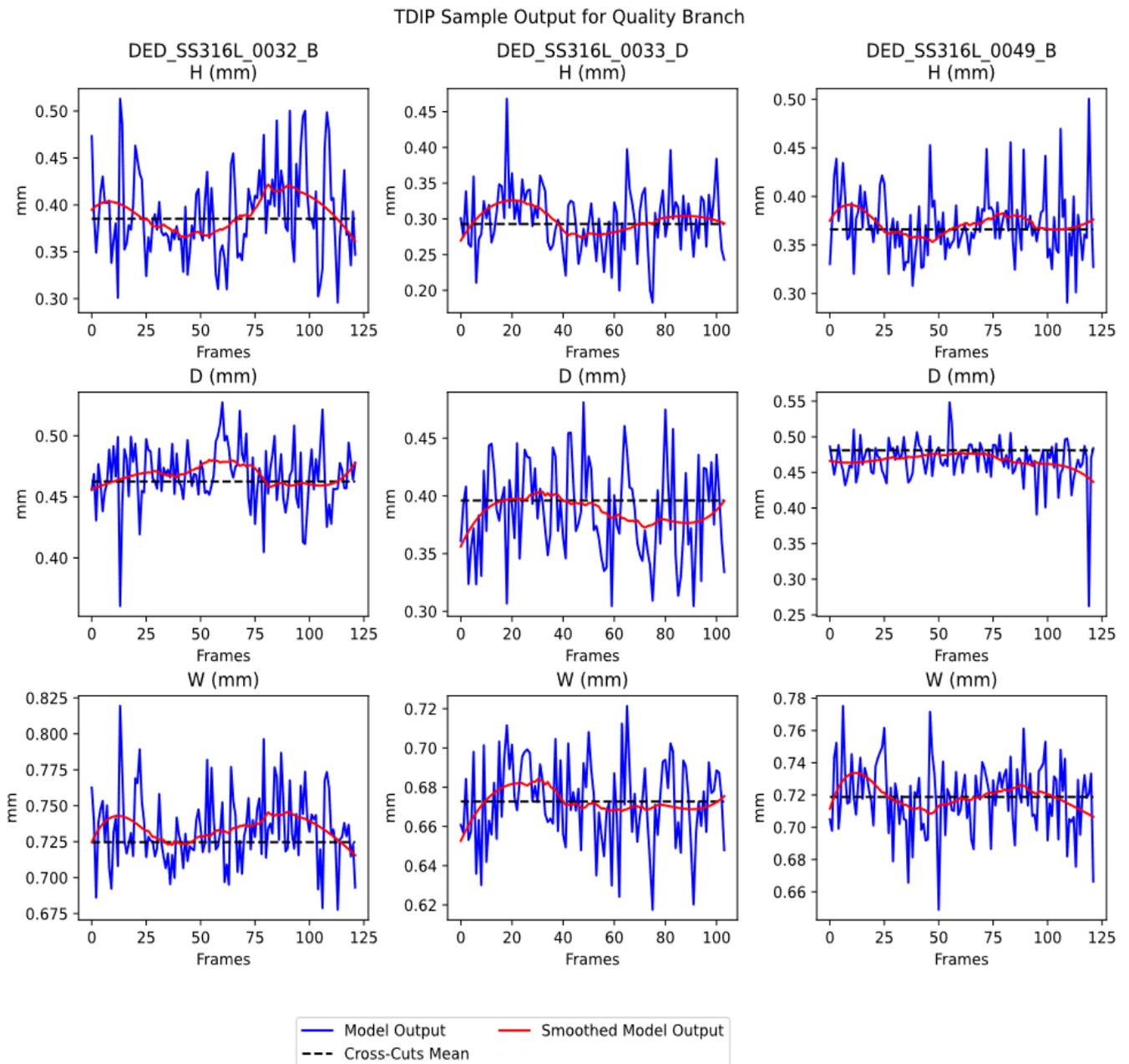

Fig. 6. Sample TDIP V.3 output for quality monitoring. This single-track process parameters are changing for each line's quarter, leading to four printing condition.

ACKNOWLEDGEMENTS

This research did not receive any specific grant from funding agencies in the public, commercial, or not-for-profit sectors.